\documentclass[10pt,journal,compsoc]{IEEEtran}
\usepackage{xcolor}

\usepackage{multirow}

\ifCLASSOPTIONcompsoc
  \usepackage[nocompress]{cite}
\else
  \usepackage{cite}
\fi
\ifCLASSINFOpdf
  \usepackage[pdftex]{graphicx}
\else
\fi
\usepackage{amsmath}
\usepackage{pifont}

\usepackage{stfloats}
\usepackage{todonotes}

\usepackage{caption}
\begin{document}
%
\title{Automated Heterogeneous Network Learning with Non-Recursive Message Passing}
%
%
%
%

\author{Zhaoqing~Li,
        Maiqi~Jiang,
        Shengyuan~Chen$^*$,
        Bo~Li,
        Guorong~Chen,
        and~Xiao~Huang
\IEEEcompsocitemizethanks{
\IEEEcompsocthanksitem Z. Li is with the Department of Systems Engineering and Engineering Management, Chinese University of Hong Kong.\protect\\
E-mail: zqli@se.cuhk.edu.hk
\IEEEcompsocthanksitem M. Jiang was with the Department of Computing, Hong Kong Polytechnic University, Hong Kong, SAR, China.\protect\\
E-mail: maiqi.jiang@connect.polyu.hk
\IEEEcompsocthanksitem S. Chen, B. Li, and X. Huang are with the Department of Computing, The Hong Kong Polytechnic University, Hong Kong, SAR, China.\protect\\
Email: \{sheng-yuan.chen, comp-bo.li, xiao.huang\}@polyu.edu.hk
\IEEEcompsocthanksitem G. Chen is with the School of Intelligent Technology and Engineering, Chongqing University of Science and Technology.\protect\\
Email: cgr@cqust.edu.cn
}
\thanks{Zhaoqing Li and Maiqi Jiang contributed equally to
this work. Corresponding author: Shengyuan Chen.}
}

%
%

\markboth{IEEE TRANSACTIONS ON KNOWLEDGE AND DATA ENGINEERING}%
{Zhao \MakeLowercase{\textit{et al.}}: Automated Heterogeneous Network Learning with Non-Recursive Message Passing}
%



\IEEEtitleabstractindextext{%
\begin{abstract}
Heterogeneous information networks (HINs) can be used to model various real-world systems. As HINs consist of multiple types of nodes, edges, and node features, it is nontrivial to directly apply graph neural network (GNN) techniques in heterogeneous cases. There are two remaining major challenges. First, homogeneous message passing in a recursive manner neglects the distinct types of nodes and edges in different hops, leading to unnecessary information mixing. This often results in the incorporation of ``noise'' from uncorrelated intermediate neighbors, thereby degrading performance. 
Second, feature learning should be handled differently for different types, which is challenging especially when the type sizes are large.
To bridge this gap, we develop a novel framework - AutoGNR, to directly utilize and automatically extract effective heterogeneous information. Instead of recursive homogeneous message passing, we introduce a non-recursive message passing mechanism for GNN to mitigate noise from uncorrelated node types in HINs. Furthermore, under the non-recursive framework, we manage to efficiently perform neural architecture search for an optimal GNN structure in a differentiable way, which can automatically define the heterogeneous paths for aggregation. Our tailored search space encompasses more effective candidates while maintaining a tractable size. Experiments show that AutoGNR
consistently outperforms state-of-the-art methods on both normal and large scale real-world HIN datasets.
\end{abstract}

}

\maketitle

\IEEEdisplaynontitleabstractindextext

%
\IEEEpeerreviewmaketitle

\IEEEraisesectionheading{\section{Introduction}\label{sec:introduction}}

%
%
%
%

\IEEEPARstart{G}{raph} neural networks (GNNs) have achieved significant success in various network learning tasks, such as node classification~\cite{nc1,nc2,nc3}, link prediction~\cite{gnn5,lp1,lp2}, and knowledge graph learning~\cite{gnnkg1,gnnkg2,gnnkg3}. However, most GNN models operate under the assumption of homogeneous graphs, which neglect the differences between various node and edge types~\cite{social1,gnn1,gnn3}. This homogeneity assumption limits the applicability of GNNs to most real-world graphs consisting of multiple types of nodes and edges—known as heterogeneous information networks (HINs)\cite{hin1,hin2}. For instance, in a citation network, nodes can represent authors and papers, which are distinct types. In HINs, interactions can occur not only between nodes of different types but also through edges that represent various types of relationships\cite{Sun2009RankClusIC}. This heterogeneity 
poses significant challenges for effectively processing and exploiting the rich structural information. Developing GNN models that can generalize across such heterogeneous structures robustly remains a major challenge~\cite{7023322}.

The limitations of existing homogeneous GNNs can be summarized in two main aspects:
$(i)$ \textbf{Noise from recursive message passing}: Traditional GNNs aggregate information from neighbors recursively~\cite{9046288}. As a result, information from irrelevant neighbors may be mixed into the final node representation, introducing noise. This noise can severely impact performance via the feature aggregation process~\cite{hetgnn}.
$(ii)$ \textbf{Lack of feature learning from node and edge types}: This limitation is particularly pronounced in heterogeneous graphs, where nodes of different types and feature spaces must be handled differently. Traditional GNNs fail to account for these distinctions, hindering the model's ability to effectively learn from the diverse features associated with various node and edge types.

Some studies aim to fully exploit the rich information in HINs using tailored GNN architectures. To mitigate the noise from message passing, several methods have been proposed to decouple the propagation and aggregation process~\cite{nrg, liu2020towards}, aggregating information from neighboring hops independently rather than recursively mixing features across hops. However, most non-recursive frameworks assume network homogeneity, failing to capture the distinct relationships in HINs and resulting in sub-optimal performance. On the other hand, to enable type-aware node representation learning, heterogeneous GNN models~\cite{rgcn, han, magnn, yang2023simple, hin2, mp1} have been developed, where meta-path-based methods~\cite{hin2, mp1} showing particular promise. Meta-paths represent higher-order relationships between nodes, transforming a HIN into sub-networks that emphasize specific relationships, enabling traditional homogeneous GNNs to be applied while mitigating unnecessary feature mixing. Despite their potential, designing suitable meta-paths can be challenging without domain-specific knowledge, and the large search space of possible meta-paths~\cite{generate_metapath} adds further complexity.

To this end, in this paper, we propose a novel method named \textbf{Auto}mated \textbf{G}NN with \textbf{N}on-\textbf{R}ecursive message passing (AutoGNR) to facilatate effective heterogeneous network learning while avoiding unnecessary information mixing.
AutoGNR is designed to automatically identify and directly utilize the heterogeneous information of HINs, performing in a search-retraining manner.
First, we introduce a non-recursive message-passing framework that enables independent aggregation of neighboring nodes at each hop. This avoids the unintentional fusion of information across hops, mitigating the risk of incorporating noise from uncorrelated neighbors. At the search stage, we introduce a tailored neural architecture search algorithm to search for a GNN structure that defines how the model aggregates information of nodes from different hops and different types. The designed search space includes both individual node types and their combinations for comprehensiveness, with a task-dependent constraint to focus on node types relevant to downstream tasks, ensuring tractability. Lastly, the retraining stage leverages the architecture parameters obtained from neural architecture search to train the model from scratch. 
We summarize our contributions as follows.

\vspace{-0.2cm}
\begin{itemize}    
    \item We propose a non-recursive framework for heterogeneous networks learning, where anchor nodes can directly utilize information (i.e., using without fusing with uncorrelated ``noise'') of neighbors of different types and hops, eliminating unnecessary information mixing.

    \item We introduce a neural architecture search algorithm tailored for heterogeneous information networks, which automatically identifies the optimal structure for aggregating information from nodes of different types and hops.
    
    \item We design a flexible search space with a task-dependent constraint, which includes more effective candidates (i.e., not only individual node types but also various combinations of them) while keeping a tractable size for efficient search.
    
    \item We conduct extensive experiments to demonstrate the effectiveness of reducing uncorrelated feature mixing, superior performance against state-of-the-art methods, and scalability on large-scale data for our proposed method.
\end{itemize}

\vspace{-0.2cm}

\section{Related Work}

\subsection{Meta-path-based Graph Neural Networks}

Heterogeneous GNN models are designed to learn complex semantic information encoded in HINs that are ignored in homogeneous models. One category of heterogeneous GNN models is to utilize meta paths to guide aggregations. HAN~\cite{han} derives sub-graphs based on manually designed meta paths and introduces a hierarchical attention mechanism, in which node-level and semantic-level attentions are used for aggregation in each meta path and to gather representations of different meta paths. MAGNN~\cite{magnn} designs several encoders to embed all intermediate nodes in meta paths, rather than just the two endpoint nodes, and incorporates different meta paths the same way as HAN. 

In addition to meta paths, methods based on meta graphs that contain richer semantic information have been developed. HAE~\cite{hae} incorporates meta paths with meta graphs and adopts node-level self-attention to capture inter-dependencies among nodes. Although these methods are able to account for different relationships in HINs, the meta structures they use need to be manually defined, a process that is extremely difficult in most cases.

\subsection{Meta-path-free Graph Neural Networks}
Another type of heterogeneous GNNs can be regarded as meta-structure-free. These methods design additional modules or mechanisms for GNNs to implicitly learn meta structures in a HIN. GTN~\cite{gtn} transforms heterogeneous networks into multiple graphs combined by adjacency matrices of different edge types weighted by attention scores and generates a new graph structure by matrix multiplications for operating GNN convolutions. HGT~\cite{hgt} introduces a node- and edge-type dependent attention mechanism to characterize the importance of each node (edge) for a target node. It also designs a sub-graph sampling algorithm to improve scalability and efficiency. MHNF~\cite{mhnf} learns a new meta-path-based graph the same way as GTN while performing aggregation on each hop separately and fusing the representations of different hops using attention. Simple-HGN~\cite{lv2021we} adopts GAT with edge-type embeddings and a two-layer recursive design to reduce complexity. SeHGNN~\cite{yang2023simple} pre-computes neighbor aggregation for efficiency, and RpHGNN~\cite{rphgnn} uses random projection to ensure linear complexity. EMGNN~\cite{emgnn} improves robustness by training a GCN on multiple graphs generated via expectation-maximization. These methods avoid the need for manually design processes. However, the recursive nature of their models and the aggregation of all meta paths still include unnecessary information mixing across different node types, leading to sub-optimal performance. 

\subsection{NAS-based Graph Neural Networks}

With the rapid development of neural architecture search (NAS) techniques, several automated network learning methods have been developed. GraphNAS~\cite{graphnas} and AGNN~\cite{agnn} use reinforcement learning to search for proper functions for each component (e.g., feature transformation, aggregation, and so on). DFG-NAS~\cite{dfgnas} searches for GNN architectures at a macro-level by determining how to integrate and organize different components. GAUSS~\cite{guass} proposes a search method for large-scale GNNs by designing an efficient lightweight supernet and introducing a joint-sampling strategy.

Recent efforts have also successfully applied NAS techniques to heterogeneous network learning problems. GEM~\cite{gem} adopts a genetic algorithm to search for meta structures and designs an attention-based multi-view GCN method to fuse heterogeneous information from different searched meta structures. AutoGEL~\cite{autogel} proposes a GNN searching framework that can explicitly utilize edge embeddings. DiffMG~\cite{diffmg} searches for a meta graph with a bi-level end-to-end optimization algorithm in a differentiable fashion. However, these methods still face problems. For example, GEM has a huge search space and the genetic algorithm that it uses makes it impossible to explore all the possibilities. DiffMG improves efficiency but facing convergence problems. Furthermore, they also struggle with unnecessary feature mixing, which not only leads to sub-optimal performance but also complicates the search process. Our method leverages NAS techniques while addressing unnecessary information mixing, improving both efficiency and performance. Unlike previous GNNs, it considers the mutual effects of node and edge types in advance, ensuring the network's semantics are captured. Additionally, we design a non-recursive message passing strategy to reduce information attenuation and simplify the search process, enabling more efficient learning in heterogeneous networks.

\vspace{-0.2cm}

\section{Preliminaries}

\textbf{Heterogeneous Information Networks.} Let $\mathcal{V},\mathcal{E},\mathcal{T},\mathcal{R}$ be the sets of node, edge, node types, and edge types, respectively. Then, a graph can be denoted by $\mathcal{G}=(\mathcal{V},\mathcal{E},\mathcal{T},\mathcal{R})$, associated with two mapping functions $\phi: \mathcal{V}\to\mathcal{T}$ and $\psi: \mathcal{E}\to\mathcal{R}$. Each node $v\in \mathcal{V}$ belongs to one type $\phi(v)\in\mathcal{T}$, and each edge $e \in \mathcal{E}$ belongs to one type $\psi(e)\in\mathcal{R}$. Based on this, a graph $\mathcal{G}=(\mathcal{V},\mathcal{E},\mathcal{T},\mathcal{R})$ is a heterogeneous information network if $|\mathcal{T}|+|\mathcal{R}|>2$. For example, in Fig.~\ref{fig1}(a), the sample HIN includes three node types: $A$, $P$, and $C$ (i.e., $|\mathcal{T}|=3$), and four edge types: $AP$, $PA$, $CP$, and $PC$ (i.e., $|\mathcal{R}|=4$).

\textbf{Graph Neural Networks.} The key idea of most GNN models is to derive representations of nodes by aggregating information of their neighbors, namely message passing mechanism, which is based on the assumption that locally connected nodes in a graph share similar properties or labels. A common message passing framework iteratively aggregates the information of neighbors for each node and combines the node's own information with the aggregated information. For an anchor node $u$, the iteration can be formalized as follows:
\begin{equation}
    \label{eq1}
    \mathbf{m}_{u}^{l}=AGG(\alpha_{uv}^{l}\mathbf{h}_{v}^{l-1}: v\in\mathcal{N}(u)),
\end{equation}
\begin{equation}
    \label{eq2}
    \mathbf{h}_{u}^{l}=UP(\mathbf{h}_{u}^{l-1},\mathbf{m}_{u}^{l}).
\end{equation}
$\mathbf{h}_{u}^{l-1}$ denotes the embedding of node $u$ at the $(l-1)$-th layer. $\mathcal{N}(u)$ denotes the set of neighbors of node $u$. $\alpha_{uv}^{l}$ is the attention score between node $u$ and $v$ that can either be a constant or be calculated by specially designed attention modules. $\mathbf{m}_{u}^{l}$ is an intermediate embedding of the information aggregated from neighbors through the $AGG$ function. Then the embedding of node $u$ can be updated through the $UP$ function by combining the embedding of itself and the intermediate embedding.

\section{Automated Heterogeneous Network Learning}
\begin{figure*}[ht]
    \centering
    \includegraphics[scale=0.43]{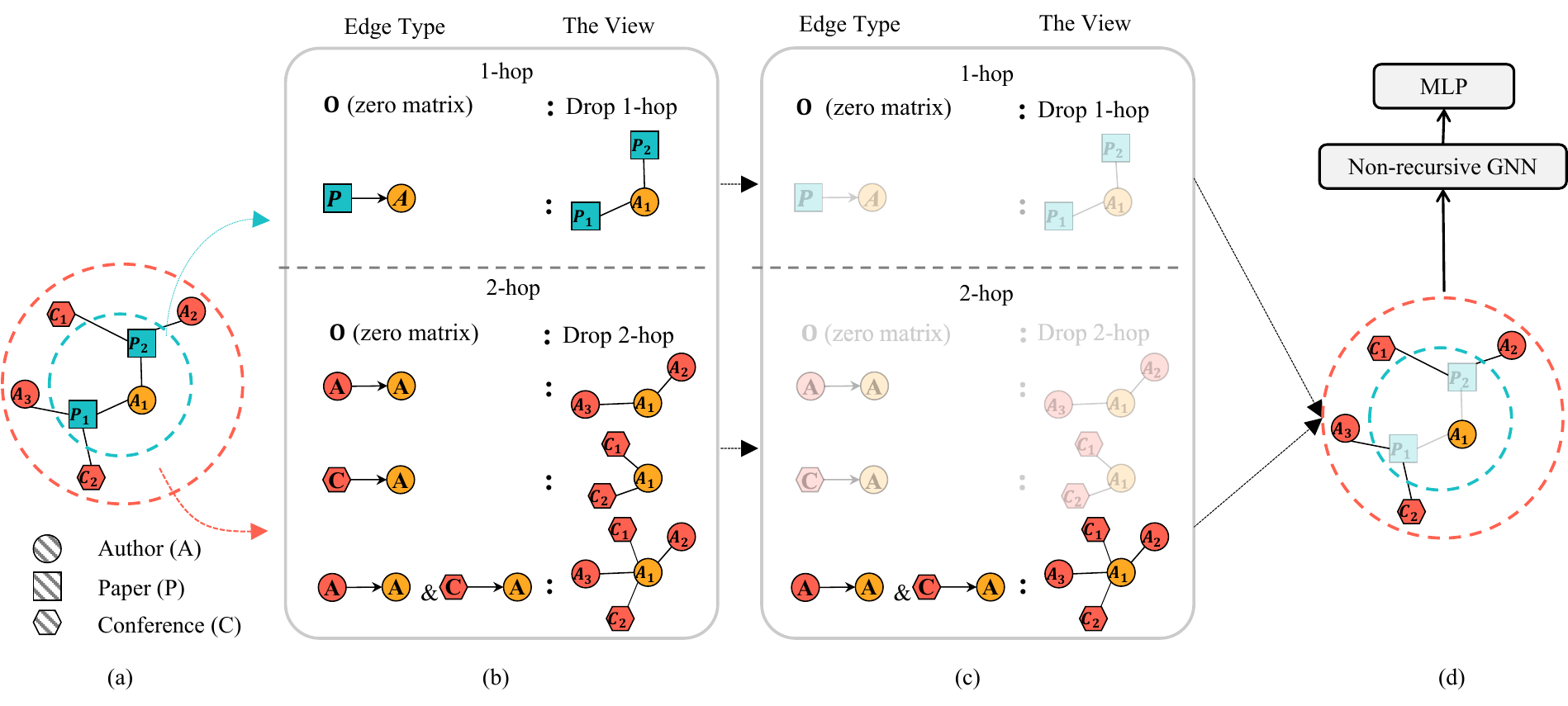}
    \vspace{-0.3cm}
    \caption{An overview of AutoGNR {with $K=2$ on an example HIN.} (a) A sample heterogeneous information network of DBLP, consisting of three node types (i.e., $A$, $P$, and $C$, denoted with different shapes) and four edge types (i.e., $AP$, $PA$, $CP$, and $PC$). We use different colors to distinguish neighbors in different hops, and here we regard the yellow Author ($A_1$) node as the anchor (target) node. (b) The specific search space for the anchor node in (a). For each hop, it includes not only individual edge-types but also the possible combinations of them. (c) The optimal architecture derived from search, which (in this example) drops out the information of Paper ($P$) neighbors in the first hop, and aggregates the information of both Paper ($P$) and Conference ($C$) nodes in the second hop. (d) Retrain the searched GNN model from scratch for downstream tasks. }
    \label{fig1}
\end{figure*}

We now introduce the proposed method. We present each component of our model in detail in the following subsections. We first introduce the homogeneous non-recursive GNN model. Then, we show our design of the heterogeneous non-recursive GNN model and how AutoGNR incorporates it into the neural architecture search framework. An overview of AutoGNR is shown in Fig.~\ref{fig1}. 

\begin{figure}[ht]
    \centering
    \includegraphics[scale=0.42]{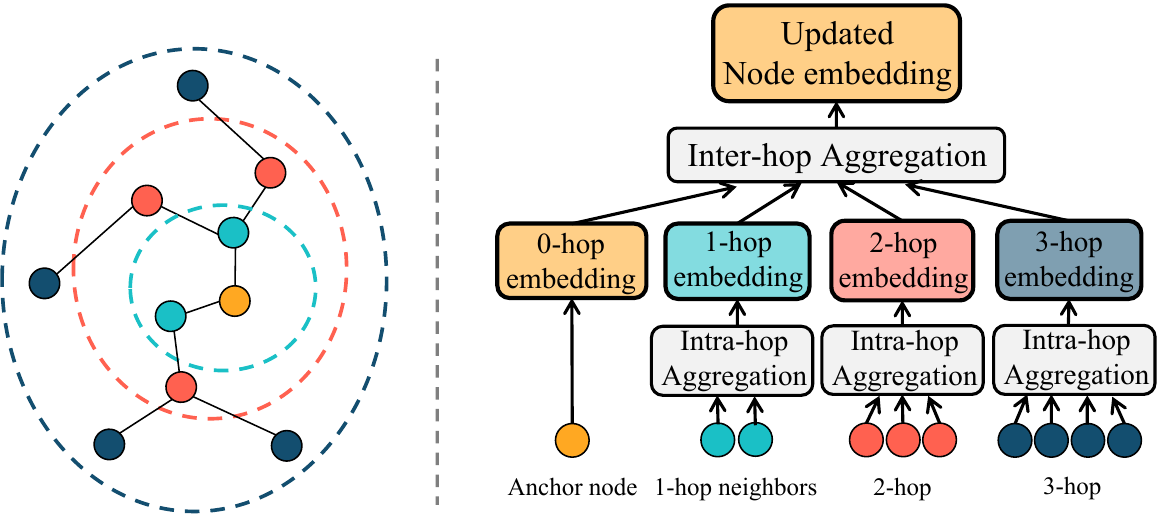}
    \caption{Illustration of non-recursive GNN framework. Different colors denote neighbors in different hops (e.g., light blue for 1-hop and red for 2-hop neighbors), with the yellow node as the anchor.}
    \label{fig2}
    \vspace{-0.5cm}
\end{figure}

\vspace{-0.2cm}
\subsection{Non-recursive Graph Neural Networks}

Most message-passing GNN models derive node representations by recursively aggregating and transforming neighbors' information across consecutive layers. This means the information from higher-order neighbors (i.e., neighbors beyond 1-hop) needs to be fused with the information of several intermediate nodes to arrive at an anchor node. However, these models are susceptible to the adverse effects of over-mixing information and noise \cite{Chen2019MeasuringAR}. This issue is exacerbated in heterogeneous network learning, where the unnecessary mixing of information across different node types, which are irrelevant to downstream tasks, results in performance degradation. For example, in an example network in Fig.~\ref{fig1}(a), the author and paper nodes have more influence on the embedding of author nodes, while the conference nodes usually cover various topics with more general features.

To eliminate unnecessary feature mixing, we first introduce a non-recursive GNN framework, which is able to directly aggregate the information of nodes in different hops. As shown in Fig.~\ref{fig2}, the non-recursive GNN consists of two stages. First, each hop's neighbors' information is aggregated separately through intra-hop aggregations to obtain intermediate hop embeddings. Then, the embedding of the anchor node is updated by combining it with the aggregated embeddings of different hops through an inter-hop aggregation. We formalize the non-recursive GNN framework as:
\begin{equation}
\label{eq5}
\begin{split}
\mathbf{e}_{u}^{k}=&AGG_{intra}(\mathbf{X};\mathcal{T})\\
= &AGG_{intra}(\alpha_{uv}\mathbf{x}_{v}: v\in\mathcal{N}^{k}(u), \phi(v)\in\mathcal{T}),
\end{split}
\end{equation}
\begin{equation}
    \label{eq6}
    \mathbf{h}_{u}=AGG_{inter}(\mathbf{x}_{u},\mathbf{e}_{u}^{1},\dots,\mathbf{e}_{u}^{K};K).
\end{equation}

$\mathbf{e}_{u}^{k}$ denotes the $k$-hop embedding for anchor node $u$. {$\mathbf{x}_{v}$ is the $v-$th row vector of $\mathbf{X}$, denoting the original feature vector of node $v$.} $\mathcal{N}^{k}(u)$ is the set of $k$-hop neighbors of node $u$. $\alpha_{uv}$ is the attention score between node $u$ and $v$, which can either be a constant or obtained by additional mechanisms. Eq.~(\ref{eq5}) means the $k$-hop embedding for node $u$ comprises all its $k-$hop neighbors regardless of their node types. As such, the node features in each hop are first aggregated separately through an intra-hop aggregation function $AGG_{intra}$. Then the embedding of the anchor node $\mathbf{h}_{u}$ can be updated by an inter-hop aggregation $AGG_{inter}$ with the embeddings of $K$ different hops, where $K$ determines the number of hops to aggregate.

\subsection{Heterogeneous Non-recursive Framework}

To include the heterogeneous information (e.g., node types), we re-formalize Eq.~(\ref{eq5}) as:
\begin{equation}
    \label{eq7}
\begin{split}
\mathbf{e}_{u}^{k}=&AGG_{intra}(\mathbf{X};\mathcal{C}^{k})\\
= &AGG_{intra}(\mathbf{x}_{v}: v\in\mathcal{N}^{k}(u), \phi(v)\in\mathcal{C}^{k}),
\end{split}
\end{equation}
where $\mathcal{C}^{k}$ is a certain subset of $\mathcal{T}$. 
In heterogeneous cases, node $u$ only aggregates information of nodes with types in $\mathcal{C}^{k}$, where the design of $\mathcal{C}^{k}$ will be discussed later. Note, for simplicity, here we omit the term of attention score.

In this work, we adopt an aggregation function as follow:
\begin{equation}
    \label{eq3}
    AGG_{intra}
    =\sum_{v\in\mathcal{N}^{k}(u), \phi(v)\in\mathcal{C}^{k}}\frac{\mathbf{x}_{v}}{\sqrt{|\mathcal{N}^{k}(u)||\mathcal{N}^{k}(v)|}}.
\end{equation}

\subsection{Tailored Search Space}
With the non-recursive framework, we can directly aggregate different hops' information without unnecessary feature mixing. However, unlike homogeneous GNNs ignoring node and edge types, in heterogeneous models, it remains necessary to determine the effective node or edge types for each hop’s information aggregation to ensure that paths correlated to downstream tasks are retained, and uncorrelated ones are discarded. As previously discussed, it is better to automatically learn a suitable structure than manual design. We apply the differentiable neural architecture search (NAS) technique to the heterogeneous non-recursive framework to enable automatic learning. 

Now we introduce how we design the search space $\mathcal{C}^{k}$ to cover more effective candidates while keeping a tractable size. Most NAS-based heterogeneous methods regard the entire node type set $\mathcal{T}$ as their search space. Usually, for each link in a GNN model, they search to select the most promising node type from $\mathcal{T}$ as the aggregating object. However, we design a more comprehensive search space $\mathcal{C}$ covering various combinations of the node types in $\mathcal{T}$ instead of only considering single node types as candidate aggregating objects. Besides, we also include a zero matrix $\mathbf{O}$, which allows the model to exclude the information from certain hops. In Fig.~\ref{fig1}(b), we illustrate the search space designed for an anchor node of a sample DBLP network (Fig.~\ref{fig1}(a)). Specifically, the search space for 1-hop neighbors $\mathcal{C}^{1}$ contains two subsets, $\mathcal{C}^{1}_{0}=\{\mathbf{O}\}$ and $\mathcal{C}^{1}_{1}=\{P\}$. Then for the 2-hop, we have $\mathcal{C}^{2}=\{\mathcal{C}^{2}_{0}=\{\mathbf{O}\}, \mathcal{C}^{2}_{1}=\{A\}, \mathcal{C}^{2}_{2}=\{C\}, \mathcal{C}^{2}_{3}=\{A,C\}\}$.

Taking advantages of the non-recursive manner, we can further design a task-dependent search space that only focuses on the heterogeneous paths related to the downstream task and exclude the unrelated ones. For example, in Fig.~\ref{fig1}, the downstream task is to predict the research field of authors, which is specifically related to the representations of author nodes ($A$). Thus, we can drop the unnecessary message passing paths related to the representations of other node types, i.e., paths from $A$ to $P$, from $A$ to $C$, and between $C$ and $P$.

We denote $|\mathcal{C}|_{max}={2}^{|\mathcal{T}|}$ as the number of all the possible node type combinations derived from $\mathcal{T}$. Then the maximum size of our search space is $(|\mathcal{C}|_{max})^{K}$. Usually, $K$ can be regarded as a small constant since the embeddings of too far neighbors would serve more as noises than effective information. (As shown later in the experiments, we find a large $K$ will lead to performance degradation). This, together with the non-recursive and task-dependent mechanisms, ensures a search space with a tractable size. Based on this, we can then perform searching in an efficient and differentiable way with an end-to-end objective function.

\subsection{Search Objective}

To make the search stage differentiable, we first introduce architecture parameters $\lambda_{c}^{k}$ for each combination of aggregating objects in $\mathcal{C}^{k}$, where $c$ is an index ranging from 0 to $|\mathcal{C}^{k}|$. Then we will mix the aggregation as follows:
\begin{equation}
    \label{eq8}
    \alpha_{c}^{k}=\frac{\operatorname{exp}(\lambda_{c}^{k})}{\sum_{c^{\prime}=0}^{|\mathcal{C}^{k}|}{\operatorname{exp}(\lambda_{c^{\prime}}^{k})}},
\end{equation}
\begin{equation}
    \label{eq9}
    \overline{AGG}(\mathbf{X};\mathcal{C}^{k})=\sum\nolimits_{c=0}^{|\mathcal{C}^{k}|}\alpha_{c}^{k}\cdot AGG_{intra}(\mathbf{X};\mathcal{C}_{c}^{k}).
\end{equation}

To this end, the mixing weights of aggregations for each hop are parameterized by a vector $\lambda^{k}$. Then the task of architecture search can be transformed into learning a set of architecture parameters $\lambda=\{\lambda^{k}\}$. Following DARTS, after the search, we can derive a discrete architecture by replacing the mixed aggregation with the most promising one, i.e.:
\begin{equation}
\begin{split}
    \label{eq10}
    AGG(\mathbf{X};\mathcal{C}^{k})&=AGG_{intra}(\mathbf{X};\mathcal{C}_{c^*}^{k}),\\ \operatorname{s.t.}\quad c^*&=\operatorname{argmax}_{c}\alpha_{c}^{k}.
\end{split}
\end{equation}
The goal of the search is to jointly learn the model parameters $\theta$ and the architecture parameters $\lambda$, since they both determine the model. Based on the NAS algorithm~\cite{anandalingam1992hierarchical, colson2007overview, darts}, this can be formalized into a bi-level optimization problem:
\begin{equation}
    \label{eq11}
    \begin{split}
        &\min\limits_{\lambda}\mathcal{L}_{valid}(\theta^{*}(\lambda),\lambda),\\
        &\operatorname{s.t.}\quad \theta^{*}(\lambda)=\operatorname{argmin}_{\theta}\mathcal{L}_{train}(\theta,\lambda),
    \end{split}
\end{equation}
where $\mathcal{L}_{valid}$ and $\mathcal{L}_{train}$ denote the validation and training loss of the downstream task.

For $AGG_{inter}$ in this paper, we use an average function to aggregate information from different neighbor hops, which is shown as:
\begin{equation}
    \label{eq13}
    \mathbf{h}_{u}=\frac{1}{K+1}(\sigma(\mathcal{F}_{norm}(\mathbf{x}_{u}))+\sum_{k=1}^{K}{\sigma(\mathcal{F}_{norm}(\mathbf{e}_{u}^{k}}))),
\end{equation}
where $\sigma(\cdot)$ denotes an activation function (e.g., GELU), and $\mathcal{F}_{norm}(\cdot)$ denotes a normalization function (e.g., $L_{2}-norm$).

\subsection{Search Algorithm}

{Now we describe how to optimize the parameters. Noting that $\theta$ and $\lambda$ are correlated with each other, we optimize the bi-level problem (i.e., Eq. (\ref{eq11})) by alternatively updating each of the two parameters while keeping the other fixed, as developed in DARTS~\cite{darts}.

\subsubsection{Updating $\theta$}
For model parameters $\theta$, they can be updated by normal gradient descent:
\begin{equation}
    \label{eq14}
    \theta=\theta-\mu_{1}\nabla_{\theta}\mathcal{L}_{train}(\theta,\lambda),
\end{equation}
where $\mu_{1}$ denotes the learning rate of $\theta$, $\nabla_{\theta}\mathcal{L}_{train}(\theta,\lambda)$ is the gradient of training loss $\mathcal{L}_{train}(\theta,\lambda)$ with respect to $\theta$. 

\subsubsection{Updating $\lambda$}
Following DARTS, we adopt the one-step unrolled learning to derive an approximate gradient of validation loss $\mathcal{L}_{valid}$ with regard to $\lambda$:
\begin{equation}
    \label{eq15}
    \nabla_{\lambda}\mathcal{L}_{valid}(\theta^{*}, \lambda)\approx\nabla_{\lambda}\mathcal{L}_{valid}(\theta-\mu_{1}\nabla_{\theta}\mathcal{L}_{train}(\theta,\lambda), \lambda),
\end{equation}
where $\theta^{*}$ is approximated by the current $\theta$ after a one-step update. 

In practice, if we set $\mu_{1}=0$ in Eq.~(\ref{eq15}), the gradient of $\lambda$ becomes $\nabla_{\lambda}\mathcal{L}_{valid}(\theta,\lambda)$, which is considered the first-order approximation. In this case, the calculation of the second-order term is eliminated, leading to a faster search. Yet, sometimes, it tends to be more difficult to converge. 
}

\begin{table}[t]
\captionsetup{justification=raggedright, singlelinecheck=false}
\begin{center}
\caption{Statistical information of datasets}
\label{table:1}
\begin{tabular}{l c c c c c} 
 \hline
 Datasets & ACM & DBLP & IMDB & PubMed  & DBLP2\\
 \hline
  \# Nodes & 8,994 & 18,405 & 12,772 & 63,109 & 1,989,077 \\
 \# Node Types & 3 & 3 & 3 & 4 & 4 \\
 \# Edges & 25,922 & 67,946 & 37,288 & 244,986 & 275,940,913 \\
 \# Edge Types & 4 & 4 & 4 & 17 & 8 \\
 \# Features & 1,902 & 334 & 1,256 & 200 & 300 \\
 \hline
\end{tabular}
\end{center}
\end{table}

\begin{table}[h]
    \captionsetup{justification=raggedright, singlelinecheck=false}
    \centering
    \caption{Comparison of AutoGNR with baseline methods. \ding{51} and \ding{55} denote whether a method faces the problem of unnecessary feature mixing or requires manual design.}
    \begin{tabular}{cccc}
        \hline
        Method & Type & Feature mixing & Manual \\
        \hline
        GAT~\cite{gnn2} & Homogeneous & - & - \\
        GCN~\cite{citation2} & Homogeneous & - & - \\
        MAGNN~\cite{magnn} & Heterogeneous & \ding{51} & \ding{51} \\
        HAN~\cite{han} & Heterogeneous & \ding{55} & \ding{51} \\
        HGT~\cite{hgt} & Heterogeneous & \ding{51} & \ding{55} \\
        RGCN~\cite{rgcn} & Heterogeneous & \ding{51} & \ding{55} \\
        Simple-HGN \cite{lv2021we} & Heterogeneous & \ding{51} & \ding{55} \\
        SeHGNN \cite{yang2023simple} & Heterogeneous & \ding{51} & \ding{55} \\
        RpHGNN \cite{rphgnn} & Heterogeneous & \ding{51} & \ding{55} \\
        EMGNN \cite{emgnn} & Both & \ding{51} & \ding{55} \\
        DiffMG~\cite{diffmg} & Heterogeneous & \ding{51} & \ding{55} \\
        \textbf{AutoGNR (Ours)} & Heterogeneous & \ding{55} & \ding{55} \\
        \hline
        
    \end{tabular}
    \label{tab:methods}
\end{table}

\section{Experiments}
In this section, we conduct experiments and provide an analysis to demonstrate the efficacy of our model. We mainly focus on the following aspects: 
\begin{enumerate}
    \item[(\textbf{A1})] the performance of AutoGNR on downstream tasks,
    \item[(\textbf{A2})] the time cost of training AutoGNR compared with that of traditional GNN models and advanced NAS-based GNN methods,
    \item[(\textbf{A3})] the capability of capturing heterogeneous information of HINs,
    \item[(\textbf{A4})] the scalability of AutoGNR on large-scale HINs.
\end{enumerate}

\begin{table*}[h]
  \centering
  \caption{Macro-F1 and Micro-F1 scores ($\pm$ standard deviation) for node classification task across different training proportions on three normal-scale heterogeneous network datasets (ACM, DBLP, and
IMDB). The results are averaged over 50 runs. The best result is highlighted in the bold, and the second-best result is underlined.}
  \vspace{-0.2cm}
  \resizebox{\linewidth}{!}{
    \begin{tabular}{ccccccccccc}
    
    \hline
          \multirow{2}{*}{Metric} &
          \multirow{2}{*}{Method} &
          \multicolumn{3}{c}{ACM} & \multicolumn{3}{c}{DBLP} & \multicolumn{3}{c}{IMDB}   \\
          \cline{3-11}
           & 
           & \multicolumn{1}{c}{25\%} & \multicolumn{1}{c}{50\%} & \multicolumn{1}{c}{100\%} & \multicolumn{1}{c}{25\%} & \multicolumn{1}{c}{50\%} & \multicolumn{1}{c}{100\%} & \multicolumn{1}{c}{25\%} & \multicolumn{1}{c}{50\%} & \multicolumn{1}{c}{100\%} \\
          \hline
    \multirow{12}{*}{\shortstack{Macro-F1}} 
    & HAN & {90.26$\pm$1.23} & {90.77$\pm$1.23} & {92.20$\pm$0.91} & 92.11$\pm$0.86 & 92.48$\pm$0.71 & 92.88$\pm$0.71 & 57.24$\pm$2.65 & 63.69$\pm$1.31 & 66.82$\pm$1.36 \\
    & MAGNN & 90.16$\pm$1.70 & 90.86$\pm$1.59 & 91.22$\pm$1.17 & {93.33$\pm$1.06} & 93.42$\pm$1.03 & 93.46$\pm$1.22 & 59.83$\pm$4.31 & 65.94$\pm$2.89 & {69.73$\pm$2.02} \\
    \cline{2-2}
    & GAT & 88.18$\pm$12.07 & 89.55$\pm$10.08 & 91.22$\pm$3.71 & 90.74$\pm$1.45 & {93.65$\pm$0.70} & {94.49$\pm$0.81} & {61.54$\pm$1.84} & 67.22$\pm$1.02 & 69.34$\pm$1.76 \\
    & GCN & 90.01$\pm$1.61 & 91.29$\pm$1.17 & 92.40$\pm$0.82 & 87.17$\pm$1.90 & 88.96$\pm$1.70 & 93.27$\pm$1.44 & 61.48$\pm$1.61 & {67.30$\pm$1.24} & 69.77$\pm$0.95 \\
    & HGT & 87.93$\pm$1.54 & 90.15$\pm$1.10 & 91.79$\pm$1.16 & 92.11$\pm$1.15 & 92.99$\pm$1.56 & 93.60$\pm$1.04 & 54.06$\pm$5.51 & 59.52$\pm$6.38 & 67.37$\pm$3.05 \\
    & RGCN & {91.32$\pm$0.84} & {91.94$\pm$0.64} & {92.73$\pm$0.44} & 93.22$\pm$0.76 & 93.27$\pm$0.78 & 93.44$\pm$0.75 & 59.67$\pm$0.89 & 66.20$\pm$0.97 & 69.58$\pm$0.59 \\
    & Simple-HGN & 83.28$\pm$12.56 & 89.14$\pm$7.54 & 90.73$\pm$6.70 & 93.72$\pm$0.90 & 93.87$\pm$0.77 & 94.19$\pm$0.68 & 60.80$\pm$1.76 & 65.40$\pm$1.26 & 69.77$\pm$1.91 \\
    & SeHGNN & 91.73$\pm$0.71 & 92.17$\pm$4.24 & {92.79$\pm$0.44} & {94.18$\pm$0.85} & \underline{94.43$\pm$0.65} & \underline{94.65$\pm$0.64} & 60.38$\pm$4.01 & 64.94$\pm$5.82 & 70.01$\pm$1.72 \\
    & RpHGNN & 91.75$\pm$0.85 & \textbf{92.23$\pm$0.76} & \textbf{93.02$\pm$0.58} & \underline{94.36$\pm$0.90} & 94.34$\pm$0.82 & 94.41$\pm$0.81 & \underline{63.69$\pm$1.56} & \underline{68.08$\pm$1.86} & \textbf{71.37$\pm$1.79} \\
    & EMGNN & \underline{91.86$\pm$0.92} & 92.16$\pm$0.78 & 92.95$\pm$0.68 & 93.55$\pm$1.26 & 94.33$\pm$0.86 & 94.49$\pm$0.80 & 54.69$\pm$7.67 & 62.16$\pm$7.37 & 68.88$\pm$4.87 \\
    \cline{2-2}
    & DiffMG & 90.81$\pm$1.16 & 91.76$\pm$1.64 & 92.58$\pm$1.26 & 92.63$\pm$4.23 & 92.74$\pm$3.88 & 94.08$\pm$2.35 & 57.01$\pm$6.41 & 63.16$\pm$3.88 & 66.94$\pm$3.97 \\
    & \textbf{AutoGNR (Ours)} &  \textbf{91.91$\pm$0.95} & \textbf{92.23$\pm$1.06} & \underline{92.99$\pm$0.93} & \textbf{94.37$\pm$1.11} & \textbf{94.51$\pm$1.19} & \textbf{94.69$\pm$1.10} & \textbf{62.52$\pm$1.77} & \textbf{68.45$\pm$1.48} & \underline{71.23$\pm$0.97} \\
    \hline
    \multirow{12}{*}{\shortstack{Micro-F1}} 
    & HAN & 90.24$\pm$1.19 & {90.76$\pm$1.18} & {92.17$\pm$0.90} & 92.73$\pm$0.62 & 93.02$\pm$0.57 & 93.40$\pm$0.57 & 64.57$\pm$3.02 & 68.59$\pm$1.42 & 71.18$\pm$1.35 \\
    & MAGNN & 90.14$\pm$1.63 & 90.85$\pm$1.56 & 91.17$\pm$1.17 & {93.89$\pm$0.79} & 93.93$\pm$0.78 & 94.01$\pm$0.94 & 65.80$\pm$2.61 & 69.90$\pm$1.65 & {72.95$\pm$1.28} \\
    \cline{2-2}
    & GAT & 88.85$\pm$9.53 & 90.08$\pm$7.88 & 92.63$\pm$3.03 & 91.44$\pm$1.30 & {94.14$\pm$0.59} & {94.89$\pm$0.69} & 65.78$\pm$2.13 & \underline{70.54$\pm$1.09} & 72.25$\pm$1.45 \\
    & GCN & 89.99$\pm$2.40 & 91.27$\pm$1.14 & 92.37$\pm$0.83 & 89.41$\pm$1.78 & 90.70$\pm$1.28 & 93.78$\pm$1.29 & {65.82$\pm$2.40} & 70.11$\pm$1.14 & 72.61$\pm$1.05 \\
    & HGT & 87.92$\pm$1.46 & 90.15$\pm$1.10 & 91.77$\pm$1.11 & 92.78$\pm$1.03 & 93.58$\pm$0.95 & 94.15$\pm$0.85 & 63.96$\pm$2.69 & 66.89$\pm$2.84 & 71.57$\pm$1.76 \\
    & RGCN & {91.29$\pm$0.81} & {91.90$\pm$0.63} & {92.68$\pm$0.45} & 93.67$\pm$6.38 & 93.68$\pm$0.81 & 93.86$\pm$0.75 & 65.46$\pm$1.87 & 69.59$\pm$1.50 & 72.31$\pm$1.10 \\
    
    & Simple-HGN & 84.44$\pm$10.30 & 89.36$\pm$6.76 & 90.83$\pm$5.93 & 94.16$\pm$0.75 & 94.28$\pm$0.68 & 94.62$\pm$0.53 & 64.64$\pm$1.69 & 68.54$\pm$1.44 & 71.99$\pm$1.61 \\
    & SeHGNN & 91.63$\pm$0.62 & 92.09$\pm$0.43 & 92.75$\pm$0.41 & 94.59$\pm$0.66 & \underline{94.84$\pm$0.55} & \underline{95.00$\pm$0.53} & 65.35$\pm$2.70 & 68.32$\pm$2.50 & 72.79$\pm$1.57 \\
    & RpHGNN & 91.75$\pm$0.82 & \underline{92.19$\pm$0.74} & \textbf{92.97$\pm$0.59} & \underline{94.79$\pm$0.81} & 94.76$\pm$0.79 & 94.81$\pm$0.74 & \underline{65.96$\pm$1.76} & 69.42$\pm$1.99 & 72.71$\pm$1.98 \\
    & EMGNN & \underline{91.86$\pm$0.84} & 92.15$\pm$0.73 & 92.90$\pm$0.68 & 94.00$\pm$1.16 & 94.72$\pm$0.79 & 94.88$\pm$0.69 & 65.27$\pm$2.64 & 69.80$\pm$2.73 & \underline{72.93$\pm$2.21} \\
    \cline{2-2}
    & DiffMG & 90.83$\pm$1.11 & 91.75$\pm$1.58 & 92.55$\pm$1.30 & 93.13$\pm$4.07 & 93.22$\pm$4.11 & 94.49$\pm$2.28 & 64.44$\pm$3.62 & 67.55$\pm$2.72 & 70.26$\pm$2.98 \\
    & \textbf{AutoGNR (Ours)} & \textbf{91.88$\pm$0.92} & \textbf{92.20$\pm$1.01} & \underline{92.96$\pm$0.86} & \textbf{94.81$\pm$0.90} & \textbf{94.94$\pm$0.96} & \textbf{95.09$\pm$0.89} & \textbf{67.25$\pm$1.77} & \textbf{71.33$\pm$1.07} & \textbf{73.67$\pm$0.95} \\
    \hline
    \end{tabular}%
    }
  \label{tab:result}%
  \vspace{-0.2cm}
\end{table*}%

\subsection{Datasets}

We evaluate AutoGNR on node classification tasks using three popular heterogeneous datasets: ACM, DBLP, and IMDB. For comparison, we use the data from~\cite{han}. The statistical details of each dataset are shown in Table~\ref{table:1}. ACM is a citation network with three node types (paper $P$, author $A$, subject $S$) and four edge types ($PA$, $AP$, $PS$, $SP$), where paper nodes are labeled by research fields. DBLP is another citation network with three node types (conference $C$, paper $P$, author $A$) and four edge types ($PA$, $AP$, $PC$, $CP$), where authors are labeled by research fields. IMDB is a movie network containing three node types (movie $M$, actor $A$, director $D$) and four edge types ($MA$, $AM$, $MD$, $DM$), where movie nodes are labeled by categories.

Additionally, we test the scalability of AutoGNR on two large-scale datasets: DBLP2 and PubMed, from \cite{HNE}. PubMed, a gene network, includes four node types (diseases $D$, genes $G$, chemicals $C$, species $S$) and seventeen edge types. DBLP2 is a citation network with four node types (author $A$, paper $P$, venue $V$, phrase $H$) and eight edge types. Full dataset details are listed in Table~\ref{table:1}.

\subsection{Baselines and Implementation Details}

We compare our model with the state-of-the-art methods, including two homogeneous models GAT~\cite{gnn2} and GCN~\cite{citation2}; two meta-path-based heterogeneous models HAN~\cite{han} and MAGNN~\cite{magnn}; six meta-structure-free heterogeneous models HGT~\cite{hgt}, RGCN~\cite{rgcn}, Simple-HGN \cite{lv2021we}, SeHGNN \cite{yang2023simple}, RpHGNN \cite{rphgnn}, EMGNN \cite{emgnn}; and a NAS-based heterogeneous model DiffMG\footnote{In order to yield consistently stable results for comparison, for DiffMG we set the search epoch as 500.}~\cite{diffmg}. The characteristics of all the baselines and AutoGNR are summarized in Table~\ref{tab:methods}.

The implementation of each baseline method follows the default setting of the original paper unless otherwise stated. Without loss of generality, we use the same set of hyperparameters for our model on all the datasets with the learning rate being 0.01 for both search and retraining processes, the dimension of the hidden layer being 64, and the dropout rate being 0.2. We set $K=3$ as the maximum number of neighbor hops (i.e., an anchor node is allowed to aggregate the information of up to 3-hop neighbors). Besides, we use an early stopping schedule with a patience of 10. All of our experiments are conducted on a single NVIDIA RTX3090 GPU.

To address the computational challenges associated with large-scale datasets, we adopt a random walk strategy to expedite the aggregation process. Specifically, we conduct 1,000 random walks with a length of 4. This approach is consistently applied across all baseline models to ensure comparability, particularly where directly handling of extensive graph structures is impractical.

\subsection{Experimental Settings}
To conduct comparisons in a fair way, we evaluate baseline methods and our model under the same setting in terms of dataset splitting and random seeds. As for datasets, we use five-fold cross validation and apply the same split of training, validation, and test sets to all the algorithms (i.e., 53.3\% for training, 26.7\% for validation, and 20\% for testing). We ensure that there are no overlapping nodes across the different folds of the test set. To evaluate these models, we run each model 10 times with 10 different random seeds for all the models and report the averaged results. Additionally, all the models are implemented using PyTorch.

\subsection{Effectiveness Analysis}

Here we follow aspect (\textbf{A1}) to evaluate the performance of baselines and AutoGNR on the node classification task. We use Macro-average and Micro-average scores as the evaluation metrics, the results of which are reported in Table~\ref{tab:result}. 

\textbf{Meta-path-based. }HAN achieves only comparable performance on ACM, and MAGNN shows a good result only on IMDB, yet they both fail to consistently yield good performance on the other datasets. This indicates the limitations and non-generality of manually designing rules to accurately capture task-dependent heterogeneous information. 

\textbf{Meta-structure-free. }
The homogeneous model GAT can yield competitive performance even over three heterogeneous methods, including HGT, RGCN, and DiffMG, on DBLP, because GAT exploits a node-level attention mechanism implicitly assigning different importance to different node types, thus, to some extent being able to account for the heterogeneous information. RGCN performs slightly better than GCN, which indicates the introduced relation-wise transformation mechanism also helps to capture the heterogeneity. It is worth noticing that HGT, which also includes attention mechanisms, performs even worse than GCN. We conjecture this is because the transformer structure introduces high complexity, leading to the overfitting problem that harms the performance.

Simple-HGN performs similarly to GAT on DBLP but inconsistently on ACM, likely due to its reliance on GAT. SeHGNN, with its simple design, achieves competitive results across datasets, indicating that non-recursive structures handle heterogeneous graphs effectively. RpHGNN, with non-recursive updates, shows strong performance, especially on ACM, though its performance on DBLP plateaus with larger training sizes, suggesting a performance ceiling. Both RpHGNN and EMGNN, which incorporates randomness, achieve comparable results.

\textbf{NAS-based. }Although DiffMG can automatically search for a meta structure, the underlying meta graph still follows a recursive manner. As we discussed before, this may include redundant and harmful feature mixing and lead to confusion during the search process since the information can fuse with noises of uncorrelated intermediate neighbors. Besides, the search algorithm adopted in DiffMG has a convergence problem, reflected in its high standard deviations in Table~\ref{tab:result}. 

Finally, AutoGNR generally outperforms the baseline models on all three datasets. This demonstrates the importance of automatically defining and directly utilizing the task-dependent heterogeneous information of HINs. Moreover, AutoGNR has only relatively small standard deviations, which indicates it can produce consistently good performance.

\begin{figure*}[th]
    \centering
    \includegraphics[scale=0.5]
    {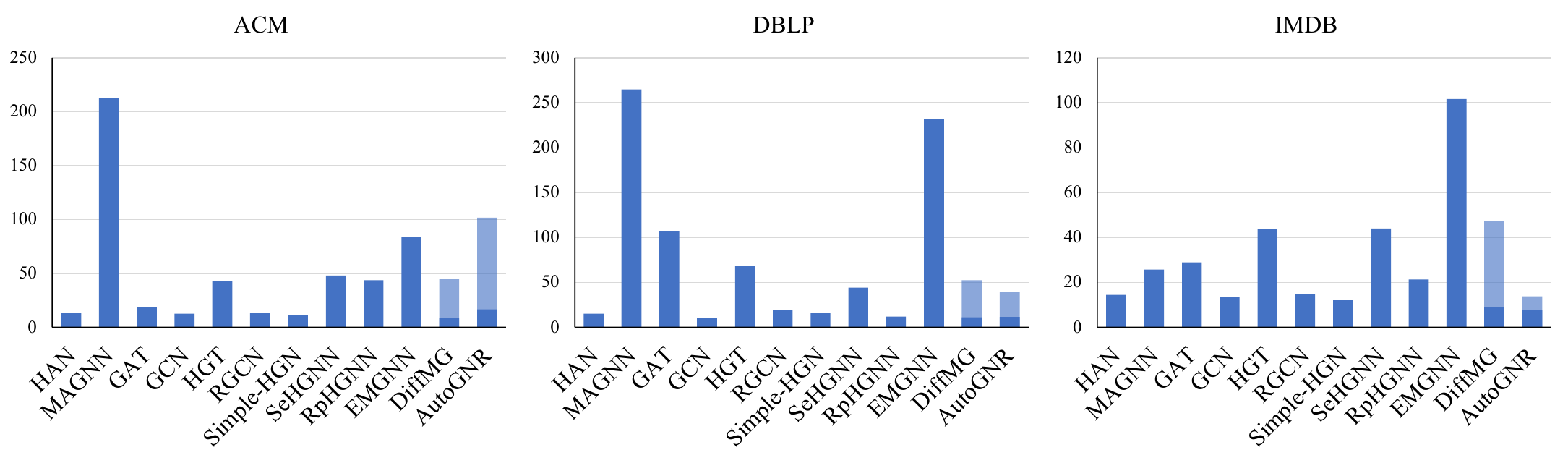}
    \vspace{-0.4cm}
    \caption{Time cost (GPU seconds) of different models on normal-scale datasets (ACM, DBLP, and
IMDB). For NAS-based models, we split the entire time cost into two parts, namely the search part (colored with light blue) and the retraining part (colored with dark blue). The results are obtained from averaging over {50} runs for each model.}
    \vspace{-0.2cm}
    \label{fig3}
\end{figure*}

\begin{figure*}[ht]
    \centering
    \includegraphics[scale=0.5]
    {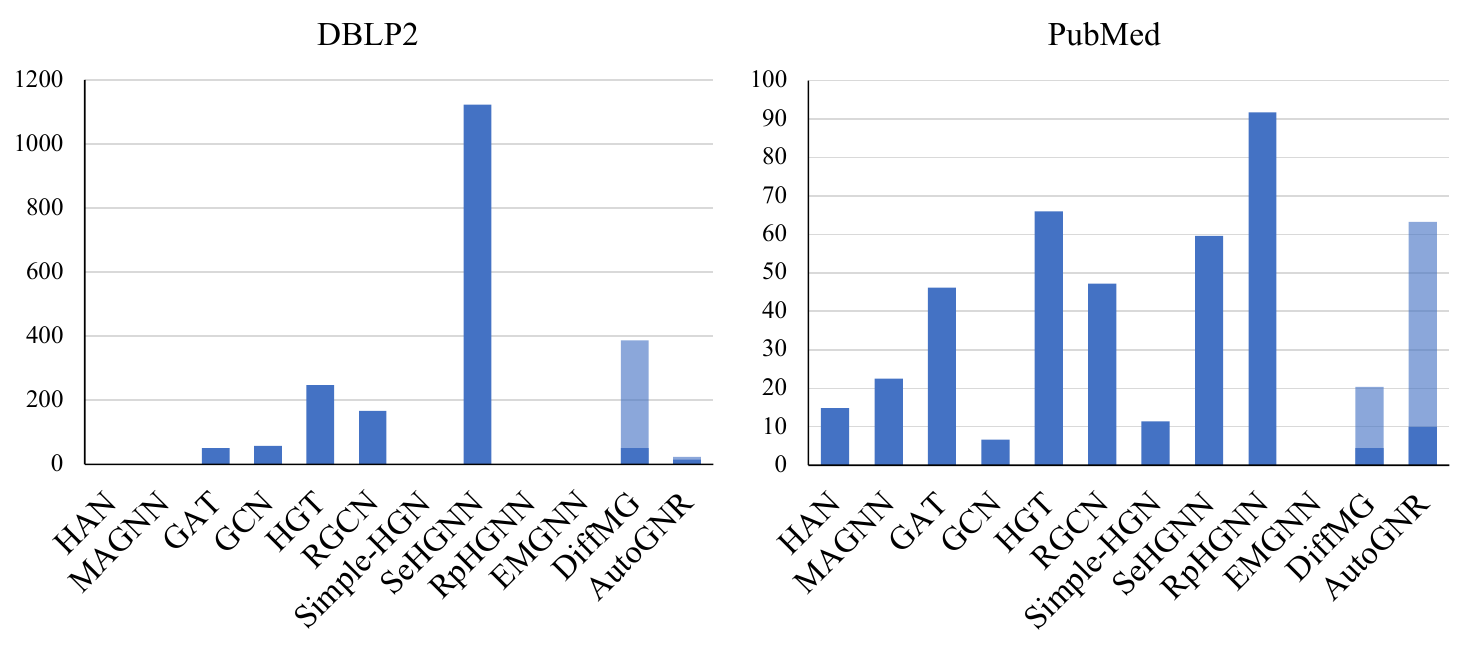}
    \vspace{-0.4cm}
    \caption{Time cost (measured in GPU seconds) of various models on large-scale datasets (DBLP2 and PubMed). For NAS-based models, the total time cost is divided into two parts: the search phase (colored light blue) and the retraining phase (colored dark blue). Models that encountered out-of-memory issues are left blank. The results represent averages over 50 runs for each model.
}
    \vspace{-0.2cm}
    \label{fig3.2}
\end{figure*}

\vspace{-0.2cm}

\subsection{Feasibility Analysis}
\label{fea}

Automated learning enables task-specific architecture search, but feasibility depends on computational costs and search space efficiency. Following aspect (\textbf{A2}), we evaluate AutoGNR's average GPU runtime compared to baselines, as shown in Fig.~\ref{fig3}, measuring the least time needed for validation loss convergence.

MAGNN incurs the highest computational cost, particularly due to its integration of RotatE. In contrast, HAN and Simple-HGN, which leverage GAT-based structures, show lower computational costs. EMGNN is inefficient for large-scale datasets, while RpHGNN performs efficiently across datasets due to its non-recursive design. SeHGNN and HGT have similar time costs, with transformer-based models contributing to higher overhead. RGCN, similar to GCN, offers modest improvements in handling heterogeneous networks but with manageable computational costs.

For the search stage, we compare AutoGNR with DiffMG which performs an efficient search by updating only one sampled operation for each link of GNN in each iteration instead of using a mixed operation. We observe that even though DiffMG performs a faster search in terms of each iteration, it needs more iterations to converge, and sometimes even fails to converge. AutoGNR, in contrast, completes the search more quickly (colored light blue) on DBLP and IMDB. In the evaluation stage, we retrain the searched model from scratch. It shows the retraining process (colored with dark blue) of AutoGNR has comparable efficiency to multiple manually-designed models. Besides, on DBLP and IMDB, even the entire two-stage process of AutoGNR can cost less time than many baseline models, demonstrating the efficiency of its search space and differentiable search algorithm.

Another noticeable issue of many heterogeneous models is their significant time cost when addressing large-scale datasets, which leads to difficulties in real-world applications \cite{lv2021we}. Here, we follow the aspects (\textbf{A2}) and (\textbf{A4}) and report the time cost of AutoGNR and the baseline models on two large-scale datasets in Fig.~\ref{fig3.2}. 

SeHGNN shows high time costs on DBLP2, exceeding 1000 seconds, despite HGT (which also uses transformers) being more efficient. We suspect SeHGNN’s aggregation of high-order neighbors through all meta paths is the cause, while HGT limits aggregation by using edge types alone. On PubMed, SeHGNN outperforms HGT, likely due to HGT’s reliance on diverse edge types. RpHGNN performs well on PubMed but has the highest time cost, suggesting its extensive use of meta paths increases computational demands. Simple-HGN, with lower performance on PubMed compared to GAT, has reduced time cost, indicating potential local minima convergence and room for improvement with better training strategies.

Similar trends can be found on AutoGNR: 1) AutoGNR is consistently feasible on large-scale datasets in terms of search and training time compared to non-automated models. 2) AutoGNR can even cost less time than non-automated models (i.e., on DBLP2), which indicates removing the redundancy of feature mixing can also benefit the training efficiency.

\vspace{-0.2cm}
\begin{figure*}[ht]
    \centering
    \includegraphics[scale=0.65]{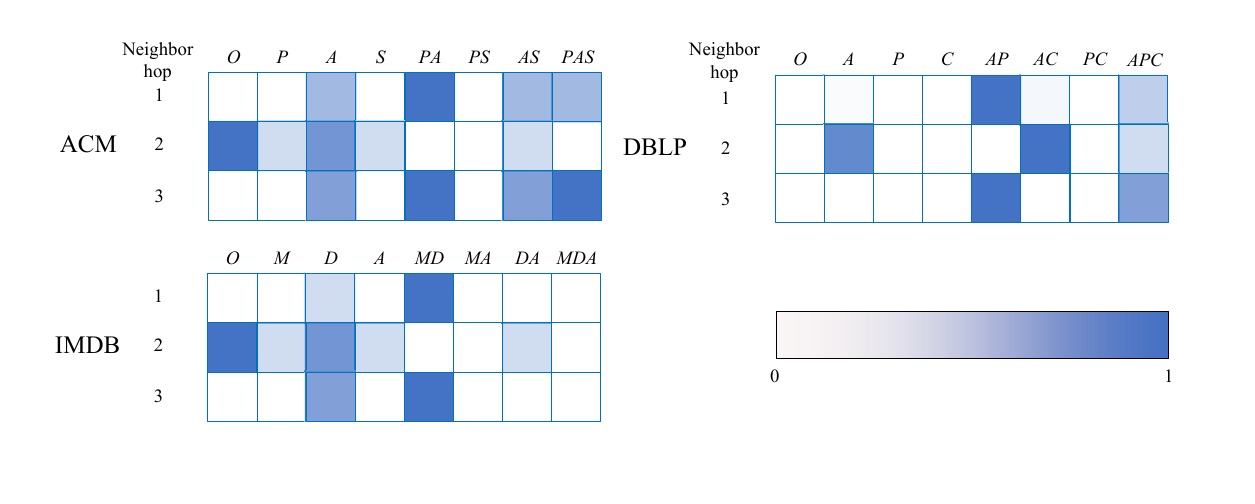}
    \vspace{-0.8cm}
    \caption{Hop-wise selection frequency distribution. The results are obtained by counting over the 50 times search on 100\% training set for each dataset. We scale the frequency to the range of [0,1] for each hop in each dataset.}
    \label{fig4}
    \vspace{-0.2cm}
\end{figure*}

\begin{table}[t]
  \centering
  \caption{The most selected architecture according to model-wise selection frequency over 50 runs.}
    \begin{tabular}{ccccc}
    \hline
    \multirow{2}{*}{Dataset} & Anchor & \multicolumn3{c}{AutoGNR} \\
    \cline{3-5}
     & Node & 1-hop & 2-hop & 3-hop \\
     \hline
    ACM & $P$ & $\{P,A\}$ & $\{A\}$ & $\{A\}$  \\
    DBLP & $A$ & $\{A,P\}$ & $\{A,C\}$ & $\{A,P\}$ \\
    IMDB & $M$ & $\{M,D\}$ & None & $\{M,D\}$ \\
    \hline
    \end{tabular}%
  \label{tab:Arch}%
  \vspace{-0.2cm}
\end{table}%

\begin{table}[t]
    \centering
    \caption{Comparison of the results ($\pm$ standard deviation) for different architectures. ``Hop-wise'' denotes that the node types for aggregation of each hop are defined independently according to the hop-wise selection frequency. ``Model-wise'' denotes the aggregation architecture defined from the Model-wise selection frequency. ``All Nodes'' denotes the architecture aggregating all node types without selection (i.e., the same as in homogeneous cases). The results are averaged over 50 runs.}
    \resizebox{0.48\textwidth}{!}{\begin{tabular}{ccccc}
    \hline
         \multicolumn{2}{c}{Method} & Hop-wise & 
         Model-wise & 
         All Nodes
         \\ \hline
         \multirow{2}{*}{ACM} & 
         Macro-F1 & $93.13\pm0.88$ & $93.55\pm0.80$ & $92.33\pm0.92$
         \\
          & 
         Micro-F1 & $93.09\pm0.79$ & $93.51\pm0.73$ & $92.30\pm0.79$
         \\ \hline
         \multirow{2}{*}{DBLP} & 
         Macro-F1 & $95.21\pm0.96$ & $95.21\pm0.96$ & $95.21\pm0.96$ 
         \\ 
          & 
         Micro-F1 & $95.57\pm0.79$ & $95.57\pm0.79$ & $95.57\pm0.79$
         \\ \hline
         \multirow{2}{*}{IMDB} & 
         Macro-F1 & $71.13\pm0.92$ & $71.13\pm0.92$ & $69.83\pm1.35$ 
         \\
          & 
         Micro-F1 & $73.55\pm0.95$ & $73.55\pm0.95$ & $72.77\pm1.17$
         \\ \hline
         
    \end{tabular}}
    \label{tab:ArchResult}
    \vspace{-0.3cm}
\end{table}

\begin{table*}[h]
  \centering
  \caption{Macro-F1 and Micro-F1 scores ($\pm$ standard deviation) for the node classification task across different training proportions on DBLP2 and PubMed. The results are averaged over 50 runs. The best result is highlighted in bold, and the second-best result is underlined.}
  \vspace{-0.2cm}
    \begin{tabular}{ccccccccccc}
    \hline
    \multirow{2}{*}{Metric} & \multirow{2}{*}{Method} & \multicolumn{3}{c}{DBLP2} & \multicolumn{3}{c}{PubMed} \\
    \cline{3-8} 
    & & 25\% & 50\% & 100\% & 25\% & 50\% & 100\% \\
    \hline
    \multirow{12}{*}{Macro-F1}
    & HAN & - & - & - & 8.82$\pm$3.22 & 10.53$\pm$3.79 & 9.83$\pm$2.86 \\
    & MAGNN & - & - & - & 5.96$\pm$2.22 & 6.13$\pm$2.19 & 5.17$\pm$1.62 \\
    \cline{2-2}
    & GAT & 9.31$\pm$3.63 & 13.00$\pm$4.08 & 16.92$\pm$4.61 & 15.57$\pm$5.93 & 27.90$\pm$6.20 & 35.28$\pm$4.56 \\
    & GCN & 6.38$\pm$2.18 & 5.72$\pm$2.04 & 5.87$\pm$1.79 & 9.14$\pm$3.36 & 10.18$\pm$3.17 & 9.28$\pm$2.78 \\
    & HGT & \underline{13.79$\pm$4.00} & 18.98$\pm$4.79 & 22.53$\pm$6.26 & 20.46$\pm$7.92 & 32.23$\pm$5.99 & 34.53$\pm$6.27 \\
    & RGCN & 4.25$\pm$2.29 & 4.04$\pm$1.52 & 4.37$\pm$1.26 & 6.98$\pm$3.00 & 7.19$\pm$2.33 & 7.61$\pm$3.11 \\
    & Simple-HGN & - & - & - & 16.30$\pm$4.09 & 19.27$\pm$6.48 & 23.09$\pm$3.18 \\
    & SeHGNN & 6.48$\pm$2.45 & 7.64$\pm$2.13 & 10.55$\pm$2.67 & \underline{23.14$\pm$4.62} & 28.88$\pm$5.12 & 35.52$\pm$3.38 \\
    & RpHGNN & - & - & - & 19.89$\pm$6.28 & 27.62$\pm$9.16 & 37.50$\pm$5.53 \\
    & EMGNN & - & - & - & - & - & - \\
    \cline{2-2}
    & DiffMG & 10.38$\pm$5.21 & \underline{22.92$\pm$5.91} & \underline{25.88$\pm$6.73} & 22.33$\pm$8.47 & \underline{36.18$\pm$6.29} & \underline{42.19$\pm$5.06} \\
    & \textbf{AutoGNR (Ours)} & \textbf{15.92$\pm$7.72} & \textbf{23.71$\pm$5.52} & \textbf{30.26$\pm$2.98} & \textbf{32.04$\pm$7.77} & \textbf{41.80$\pm$4.90} & \textbf{47.03$\pm$3.95} \\
    \hline
    \multirow{12}{*}{Micro-F1}
    & HAN & - & - & - & 15.62$\pm$4.86 & 21.60$\pm$5.15 & 22.78$\pm$4.62 \\
    & MAGNN & - & - & - & 16.96$\pm$4.51 & 18.91$\pm$4.23 & 20.71$\pm$4.16 \\
    \cline{2-2}
    & GAT & 26.47$\pm$7.55 & 30.44$\pm$6.25 & 34.10$\pm$5.10 & 22.11$\pm$6.97 & 33.93$\pm$6.69 & 39.76$\pm$4.47 \\
    & GCN & 17.17$\pm$4.98 & 16.54$\pm$4.40 & 17.98$\pm$4.54 & 13.87$\pm$5.13 & 15.80$\pm$3.77 & 15.80$\pm$3.64 \\
    & HGT & \underline{30.62$\pm$6.93} & 35.07$\pm$5.62 & 40.15$\pm$5.48 & 27.11$\pm$6.91 & 36.89$\pm$5.60 & 39.96$\pm$5.12 \\
    & RGCN & 16.98$\pm$3.79 & 17.80$\pm$3.42 & 15.85$\pm$3.36 & 15.40$\pm$5.53 & 14.09$\pm$2.39 & 20.38$\pm$3.91 \\
    & Simple-HGN & - & - & - & 21.80$\pm$5.09 & 24.87$\pm$6.11 & 27.00$\pm$3.84 \\
    & SeHGNN & 19.37$\pm$6.00 & 22.14$\pm$5.45 & 27.14$\pm$4.58 & 24.49$\pm$4.46 & 30.47$\pm$4.64 & 36.62$\pm$3.53 \\
    & RpHGNN & - & - & - & 25.91$\pm$5.81 & 32.76$\pm$7.40 & 42.09$\pm$5.03 \\
    & EMGNN & - & - & - & - & - & - \\
    \cline{2-2}
    & DiffMG & 26.89$\pm$9.59 & \underline{39.89$\pm$6.60} & \underline{43.09$\pm$6.39} & \underline{30.18$\pm$8.22} & \underline{42.36$\pm$6.73} & \underline{48.13$\pm$4.75} \\
    & \textbf{AutoGNR (Ours)} & \textbf{31.84$\pm$10.19} & \textbf{40.63$\pm$4.42} & \textbf{46.68$\pm$3.33} & \textbf{40.33$\pm$7.22} & \textbf{46.44$\pm$5.35} & \textbf{52.04$\pm$4.02} \\
    \hline
    \end{tabular}%
  \label{tab:result2}%
  \vspace{-0.2cm}
\end{table*}%

\subsection{Learned Heterogeneous Information Analysis}

Next, following the aspect (\textbf{A3}), we show the specific heterogeneous paths AutoGNR has derived from HINs to demonstrate its capability of capturing heterogeneous information. In Fig.~\ref{fig4}, we report the hop-wise selection frequency distribution\footnote{The hop-wise selection frequency is obtained by counting and averaging the search results for each hop independently. By contrast, the model-wise selection frequency is obtained by averaging the search results of the entire GNN structure including all hops.} of all the datasets, which directly implies the aggregation preference we searched on 100\% training set for each hop. The frequencies reveal a very clear preference of a certain node type combination for each hop. For example, on DBLP, the search results show an obvious preference of aggregating nodes of type $\{A,P\}$, $\{A,C\}$, and $\{A,P\}$ in 1-, 2-, and 3-hop, respectively. Moreover, even less-selected combinations have a major subset overlapping with the most preferred one. This demonstrates that AutoGNR can indeed learn task-dependent heterogeneous information from HINs. Besides, we also report the most selected architecture according to a model-wise selection frequency in Table~\ref{tab:Arch} for comparison, which shows a similar preference as hop-wise results. It can be observed the searched architecture also overlaps with some of the meta paths that have been manually developed in HAN and MAGNN (e.g., meta paths $APA$ on DBLP and $MDM$ on IMDB adopted in both HAN and MAGNN).

Then we evaluate the efficacy of the learned paths. In Table~\ref{tab:ArchResult}, we compare the results of the architectures defined by hop-wise and model-wise selection frequency with that of the architecture aggregating all node types without any selection. The results show that architectures defined by the hop-wise and model-wise frequency perform alike. They all achieve much better performance than the ``all-nodes'' architecture on ACM and IMDB, which indicates for a HIN, not all the information is necessary for a specific task, and some uncorrelated ``noises'' can even harm the performance. This again demonstrates the importance and superior capability of AutoGNR to automatically derive task-dependent heterogeneous information of HINs and drop the ineffective noises.

\subsection{Scalability Analysis}

While section~\ref{fea} explores AutoGNR's time cost on large-scale datasets, performance also defines scalability. Following aspect (\textbf{A4}), we assess AutoGNR's performance on PubMed and DBLP2 (Table~\ref{tab:result2}, "-" indicates memory overflow).

\textbf{Meta-path-based Approaches.} Both HAN and MAGNN could not be executed on DBLP2 due to excessive memory demands resulting from their meta paths and recursive structures. In the case of PubMed, their performances were underwhelming, consistently inferior to that of GCN. This limitation underscores the ineffectiveness of manually crafted meta paths in managing the complexities inherent in expansive graph structures, likely due to an inability to adapt domain-specific knowledge to extensive, intricate networks.

\textbf{Meta-structure-free Approaches.} GGAT outperformed GCN on both datasets, demonstrating the effectiveness of its attention mechanism in capturing heterogeneous information. Contrarily, RGCN exhibited slightly poorer performance than GCN, possibly because it treats all relations uniformly, making it challenging to discern useful information amidst a high volume of non-contributory relations. Notably, HGT showed improved outcomes on the larger datasets compared to smaller ones, suggesting that its transformer architecture is well-suited to managing high complexity and diversity of edge types, thereby mitigating issues related to overfitting.

Simple-HGN, RpHGNN, and EMGNN were unable to execute on DBLP2 due to high memory demands caused by their recursive or iterative structures. Moreover, EMGNN also failed to execute on PubMed, indicating that its graph generation step is especially resource-intensive. While Simple-HGN, SeHGNN, and RpHGNN performed reasonably on PubMed, their results fell short of HGT’s performance. This gap suggests that although these models are designed for efficiency, their structures may not scale effectively to accommodate the complexities of larger heterogeneous graphs.

\textbf{NAS-based Approaches.} DiffMG consistently secured near-top performances in these datasets, yet it continued to experience convergence issues, as evidenced by its high standard deviations. 

Finally, AutoGNR consistently achieves the best performance on both datasets, reinforcing its superiority and affirming its capacity to handle large-scale and complex heterogeneous graph scenarios. This further underscores the advantages of AutoGNR in automatically defining and leveraging task-dependent heterogeneous information in HINs while eliminating unnecessary feature mixing.

\begin{figure*}[ht]
    \centering
    \includegraphics[scale=0.40]{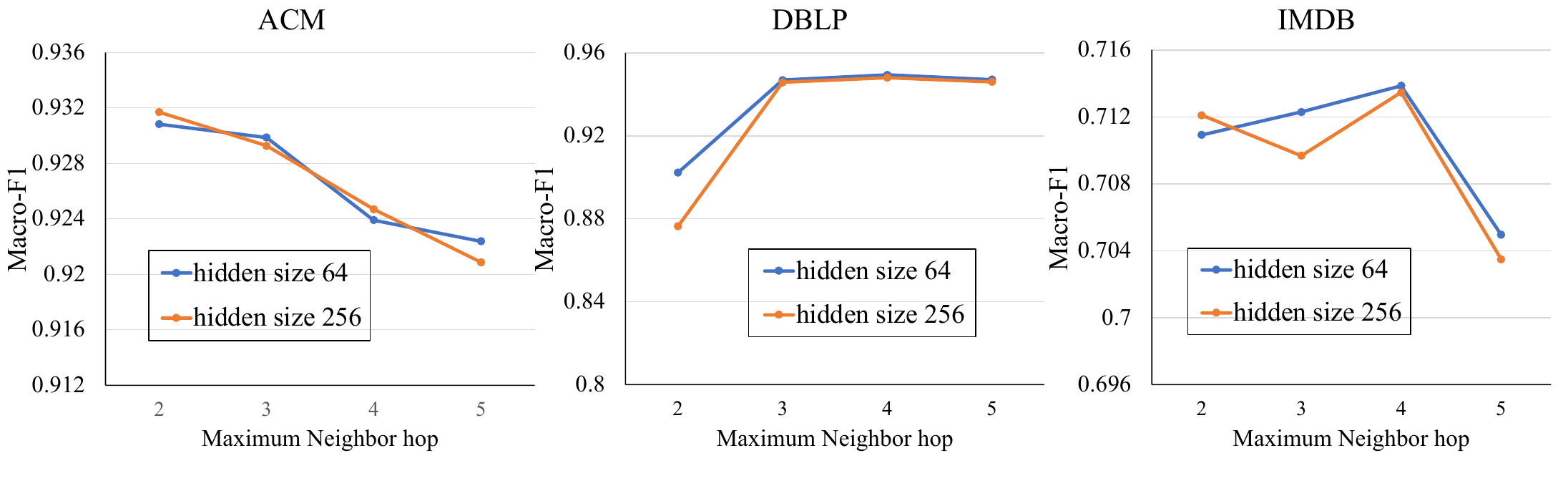}
    \vspace{-0.2cm}
    \caption{Macro-F1 performance of AutoGNR with different hyperparameters (i.e., hidden dimension width and maximum number of neighbor hops) over 50 runs on 100\% training set.}
    \label{fig5}
    \vspace{-0.2cm}
\end{figure*}
\begin{table*}[ht]
  \centering
  \caption{Ablation Study Results ($\pm$ standard deviation). The results are averaged over 50 runs. The best results are highlighted in bold, and the second best are underlined.}
    \begin{tabular}{ccccccc}
    \hline
          \multirow{2}{*}{Method} & \multicolumn{2}{c}{ACM}        & \multicolumn{2}{c}{DBLP}        & \multicolumn{2}{c}{IMDB}   \\
          \cline{2-3} \cline{4-5} \cline{6-7}
           & \multicolumn{1}{c}{Macro-F1} & \multicolumn{1}{c}{Micro-F1} & \multicolumn{1}{c}{Macro-F1} & \multicolumn{1}{c}{Micro-F1} & \multicolumn{1}{c}{Macro-F1} & \multicolumn{1}{c}{Micro-F1} \\
          \hline
    Supernet &
    $91.15\pm0.89$ & $91.11\pm0.84$ & $94.27\pm0.81$ & $94.68\pm0.60$ & $69.64\pm2.02$ & $72.41\pm1.81$ \\
    Weight-transfer & $92.48\pm1.16$ & $92.45\pm1.17$ & $94.25\pm1.13$ & $94.64\pm0.95$ & $69.72\pm1.75$ & $72.23\pm1.38$ \\
    Fine-tuning &  $92.87\pm0.83$ & $92.83\pm0.78$ & $94.48\pm1.07$ & $94.86\pm0.86$ & $70.82\pm1.06$ & $73.31\pm1.08$ \\
    \cline{1-1}
    AutoGNR (DiffMG) &  92.04 $\pm$ 1.68 &  92.01 $\pm$ 1.64 & 90.00 $\pm$ 6.08 & 90.58 $\pm$ 5.84 & 69.60 $\pm$ 3.91 & 72.35 $\pm$ 3.16 \\    
    AutoGNR (First Order) &  \textbf{92.99 $\pm$ 0.93} &  \textbf{92.96 $\pm$ 0.86} & \textbf{94.72 $\pm$ 1.22} & \textbf{95.11 $\pm$ 1.00} & \textbf{71.23 $\pm$ 0.97} & \textbf{73.67 $\pm$ 0.96} \\
    \cline{1-1}
    AutoGNR &  \textbf{92.99 $\pm$ 0.93} &  \textbf{92.96 $\pm$ 0.86} & \underline{94.69 $\pm$ 1.10} & \underline{95.09 $\pm$ 0.89} & \textbf{71.23 $\pm$ 0.97} & \textbf{73.67 $\pm$ 0.96} \\
    \hline
    \end{tabular}%
  \label{tab:finetune}%
\end{table*}%

\subsection{Ablation Study}

In this part, we investigate the effects of the search-retrain manner and search algorithm on AutoGNR. Recall that AutoGNR retrains the searched model for final evaluations. In Table~\ref{tab:finetune}, we report the results of several methods without retraining the model from scratch, namely Supernet, Weight-transfer, and Fine-tuning. We refer to the mixed-operation model (i.e., including all the candidate links at the search stage) as the supernet. To avoid retraining, we can directly evaluate the performance of the trained supernet on each dataset since AutoGNR searches for architectures with an end-to-end optimization. Apart from mixed-operation, another way is to use the searched discrete architecture with the weight parameters directly transferred from the corresponding weights of a trained supernet, which we refer to as Weight-transfer. Besides, we can also apply fine-tuning to the Weight-transfer model, which only requires a few more weight updates. We find that the supernet with mixed-operation performs the worst, because it fuses with too much unnecessary information that may harm the performance. By contrast, the discrete Weight-transfer model derived from the supernet performs much better, and can be even better with fine-tuning, which indicates dropping out uncorrelated information benefits the model performance a lot. Although the discrete models have achieved comparable results, we find AutoGNR with a retraining process can still outperform all of these methods, which explains why we adopt a search-retrain manner.

We then study two different search algorithms. One is the same as the search algorithm of DiffMG, and the other is the first-order approximation as discussed in the Search Algorithm section. Although these two algorithms can further accelerate the search stage for AutoGNR, they cannot consistently yield good performance, which is reflected in their high standard deviations. This validates the rationality of AutoGNR to adopt the one-step unrolled search algorithm, which achieves a proper balance between efficiency and performance.

Finally, in Fig.~\ref{fig5}, we explore the effect of the maximum number of neighbor hops and the hidden dimension width on AutoGNR. Several trends can be observed. First, the hidden dimension width has no significant influence on the performance on all three datasets. Second, as the maximum neighbor hop number increases to 4, the performance shows a general improvement on DBLP and IMDB and a degradation on ACM, while as it further increases to 5, the performance starts decreasing on all three datasets. This indicates that a larger maximum neighbor hop number may lead to a larger search space including more possibilities. However, when it gets too large, farther neighbors may contain more uncorrelated noise than effective information, which will harm the representational ability of the model and reduce the final performance. Hence, in this paper, we choose a moderate maximum neighbor hop number (i.e., $K=3$) to conduct our experiments for all three datasets.

\section{Conclusion}

This work addresses key challenges in heterogeneous network learning by proposing a non-recursive GNN framework. The framework directly aggregates information from nodes of different types or hops, avoiding noise from uncorrelated intermediates. We also design a tailored search space to enable efficient discovery of optimal GNN structures in a differentiable manner. Our approach is compatible with any GNN model based on the message-passing framework, making it versatile and beneficial for future studies. Extensive experiments show that our method reduces irrelevant information mixing, improves training efficiency, and achieves superior performance on HINs compared to state-of-the-art models. Future work will focus on combining search and training into a single stage to further enhance scalability for large networks.

\vspace{-0.1cm}

\bibliography{ref}



%

\ifCLASSOPTIONcaptionsoff
  \newpage
\fi



\bibliographystyle{IEEEtran}

%



%

\begin{IEEEbiography}[{\includegraphics[width=1in,height=1.25in,clip,keepaspectratio]{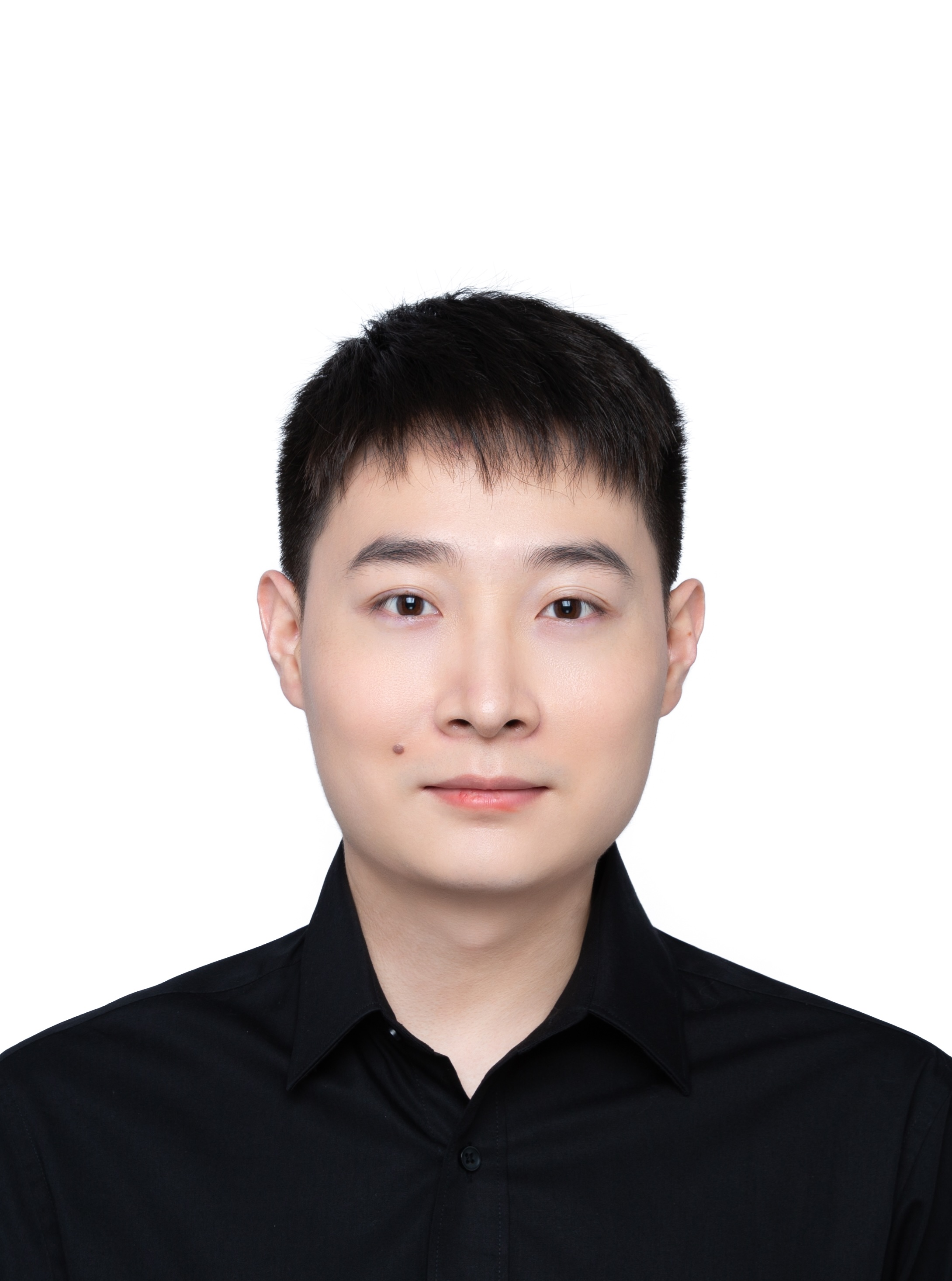}}]{Zhaoqing Li}
received the B.S and M.S. degree in automation from Northwestern Polytechnical University, China. He is currently pursuing the Ph.D. degree with the Chinese University of Hong Kong, Hong Kong SAR, China. His current research interests include machine learning and speech recognition with a special focus on developing lightweight and efficient deep learning systems.
\end{IEEEbiography}

\begin{IEEEbiography}[{\includegraphics[width=1in,height=1.25in,clip,keepaspectratio]{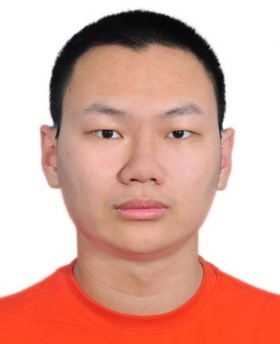}}]{Maiqi Jiang}
received the B.S. degree from the College of Medicine and Biological Information Engineering, Northeastern University, China, in 2021, and the M.S. degree from the Department of Computing, The Hong Kong Polytechnic University, Hong Kong, in 2023. His research interests include graph neural network, social network analysis, and knowledge graph.
\end{IEEEbiography}

\begin{IEEEbiography}[{\includegraphics[width=1in,height=1.25in,clip,keepaspectratio]{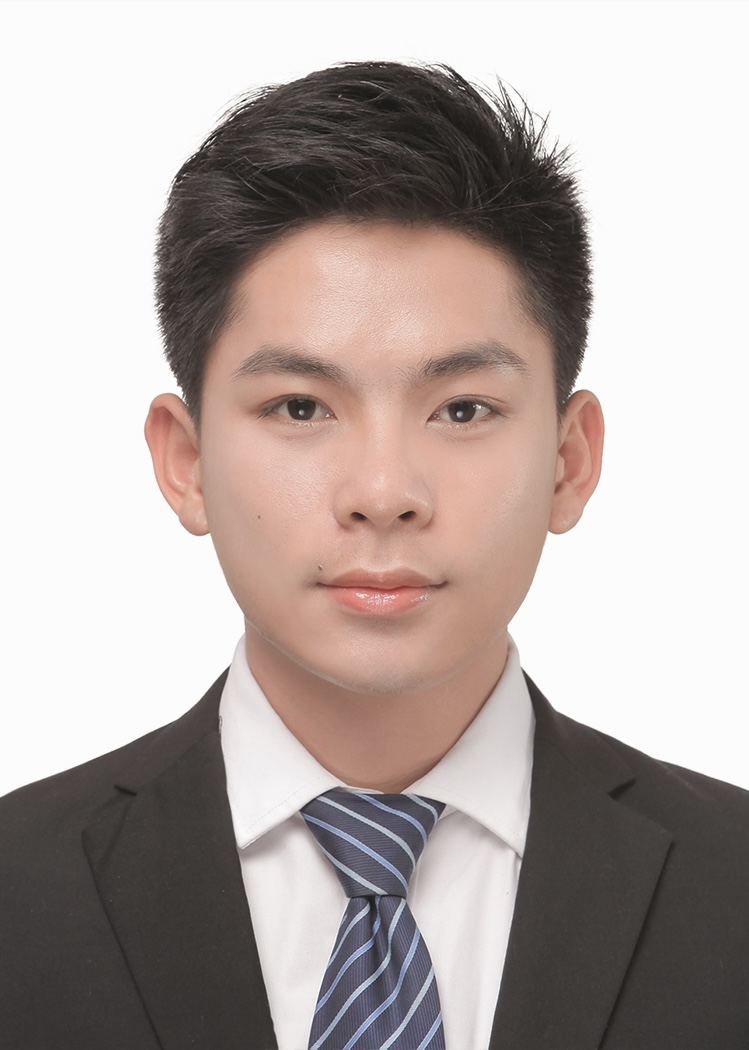}}]{Shengyuan Chen}
is a Postdoctoral Fellow in the Department of Computing at The Hong Kong Polytechnic University. He received his Ph.D. degree from the same department and his B.S. from Fudan University. His research interests include reasoning, large language models, and knowledge graphs.
\end{IEEEbiography}

\begin{IEEEbiography}[{\includegraphics[width=1in,height=1.25in,clip,keepaspectratio]{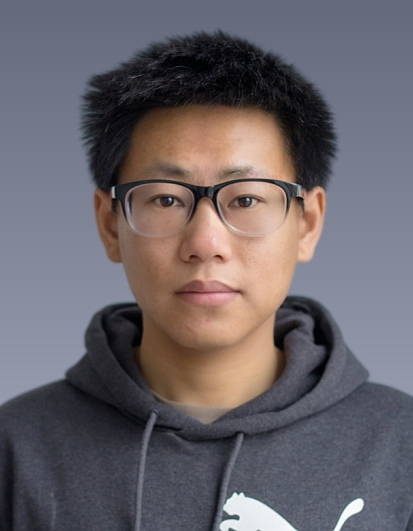}}]{Bo Li}
is an assistant professor in the Department of Computing at The Hong Kong Polytechnic University. Formerly, he was a Postdoctoral Fellow at the University of Oxford and the University of Texas at Austin. He received his Ph.D. in Computer Science from Stony Brook University and B.S. in Applied Maths from Ocean University of China. He is broadly interested in algorithms, AI, and game theory.
\end{IEEEbiography}

\begin{IEEEbiography}[{\includegraphics[width=1in,height=1.25in,clip,keepaspectratio]{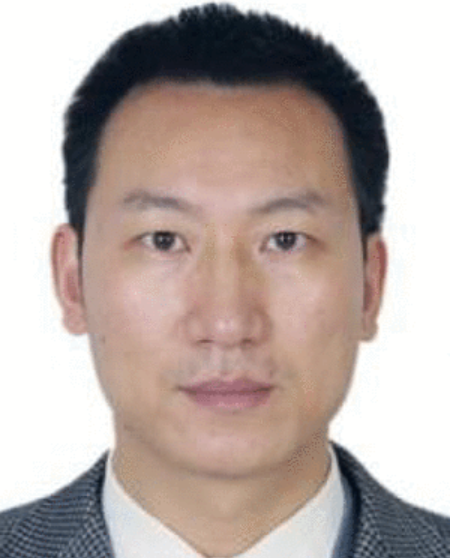}}]{Guorong Chen}
is a professor in School of Intelligent Technology and Engineering at Chongqing University of Science and Technology. He received Ph.D. in Mechanical Manufacturing and Automation from Chongqing University in 2011, M.S. in Software Engineering from Chongqing University in 2004, and B.S. in Automation from Nanjing University of Technology in 1998. His research interests include image processing, complex networks, knowledge graphs. He is a chair of ICAICA2024.
\end{IEEEbiography}
\vspace{-5.5in}

\begin{IEEEbiography}
[{\includegraphics[width=1in,height=1.25in,clip,keepaspectratio]{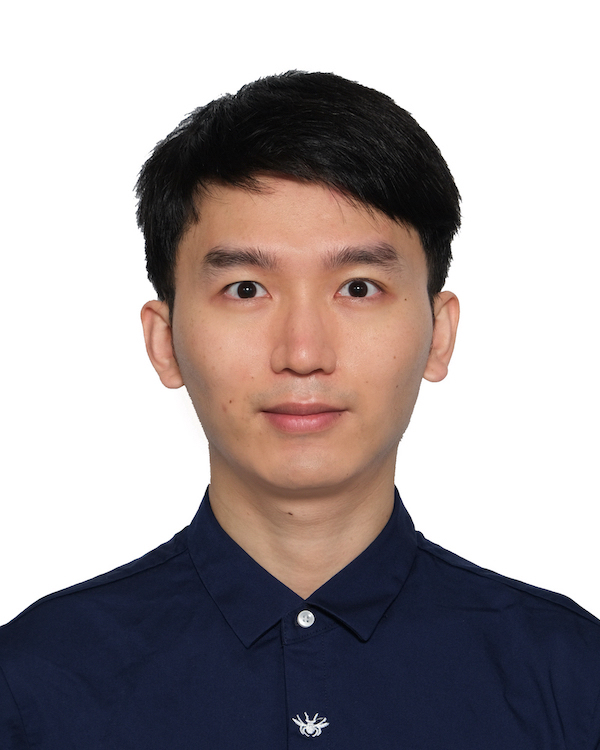}}]{Xiao Huang}
is an assistant professor in the Department of Computing at The Hong Kong Polytechnic University (PolyU). He is an associate director of the Collaborative Generative Al (Co-GenAl) Research Centre at PolyU. He received a Ph.D. in Computer Engineering from Texas A\&M University in 2020, an M.S. in Electrical Engineering from Illinois Institute of Technology in 2015, and B.S. in Engineering from Shanghai Jiao Tong University in 2012. His research interests include graph representation learning, knowledge graphs, retrieval-augmented generation, recommender systems, and text-to-SQL. He won the Best Paper Award Honorable Mention at SIGIR 2023. He is a PhD Symposium Chair of ICDE 2025. Before joining PolyU, he worked as a research intern at Microsoft Research and Baidu USA.
\end{IEEEbiography}







\end{document}